# Cross Feature Selection to Eliminate Spurious Interactions and Single Feature Dominance Explainable Boosting Machines⁻

Shree Charran R[1]

*Indian Institute of Science, Bangalore, India*

Sandipan Das Mahapatra[2]

*Indian Institute of Technology, Kanpur, India*

*Abstract* – Interpretability is a crucial aspect of machine learning models that enables humans to understand and trust the decision-making process of these models. In many real-world applications, the interpretability of models is essential for legal, ethical, and practical reasons. For instance, in the banking domain, interpretability is critical for lenders and borrowers to understand the reasoning behind the acceptance or rejection of loan applications as per fair lending laws. However, achieving interpretability in machine learning models is challenging, especially for complex high-performance models. Hence Explainable Boosting Machines (EBMs) have been gaining popularity due to their interpretable and high-performance nature in various prediction tasks. However, these models can suffer from issues such as spurious interactions with redundant features and single-feature dominance across all interactions, which can affect the interpretability and reliability of the model's predictions. In this paper, we explore novel approaches to address these issues by utilizing alternate Cross-feature selection, ensemble features and model configuration alteration techniques. Our approach involves a multi-step feature selection procedure that selects a set of candidate features, ensemble features and then benchmark the same using the EBM model. We evaluate our method on three benchmark datasets and show that the alternate techniques outperform vanilla EBM methods, while providing better interpretability and feature selection stability, and improving the model's predictive performance. Moreover, we show that our approach can identify meaningful interactions and reduce the dominance of single features in the model's predictions, leading to more reliable and interpretable models.

## I. INTRODUCTION

Machine learning models have become ubiquitous in various domains, including healthcare, finance, and law. These models have shown remarkable success in solving complex tasks, such as image recognition, natural language processing, and predictive analytics. However, as these models become more complex and sophisticated, their interpretability and Explainability become more challenging. Interpretability refers to the ability of humans to understand how a model arrives at its predictions or decisions. Explainability, on the other hand, refers to the ability to explain the model's internal workings in a clear and concise manner.

The importance of interpretability and Explainability has been widely recognized in various domains, including healthcare, finance, and law. In healthcare, interpretability is critical for doctors and patients to understand the reasoning behind the model's predictions and to make informed decisions. In finance, interpretability is necessary for regulatory compliance, risk management, and fraud detection. In law, interpretability is essential for legal decision-making, where the reasoning behind the decision needs to be explained to the stakeholders.

Despite the importance of interpretability and Explainability, achieving these goals in machine learning models is challenging, especially for complex models such as deep neural networks and tree-based boosting models. These models can generate accurate predictions, but their internal workings can be difficult to interpret and understand. Hence Explainable Boosting Machines (EBMs) have been gaining popularity as EBMs are a glass box model, designed to have accuracy comparable to state-of-the-art machine learning methods like Boosted Trees, while being highly intelligible and explainable. Despite EBM bringing the best of both performance and interpretability, it suffers from Spurious interactions with redundant features and single feature dominance in interactions.

Spurious interactions occur when the model predicts a mathematical relationship in which two or more variables are associated but not causally related, due to either coincidence or usually one of the features in the interaction are redundant noise and have no/low feature importance individually. But somehow a feature interacts to form high feature importance scores.

Single-feature dominance occurs when a single feature dominates the model's predictions and interactions, making it difficult to understand the contribution of other features. Usually, one single feature which is highly correlated to multiple features occurs in multiple interactions leading to non-meaningful interpretations. It is a known that all interactions need not be meaningful and score high in feature importance and vice versa, and it can also be said redundancy is a tricky matter to judge. Thus, these issues can lead to models that are difficult to interpret and trust, limiting their applicability in real-world domains.

Various methods have been proposed to enhance the interpretability and transparency of machine learning models. These methods can be broadly categorized into two categories: model-specific and model-agnostic methods. Model-specific methods focus on developing specialized algorithms that are designed to generate interpretable models. For example, decision trees are a model-specific method that generates a set of rules that can be easily understood by humans. However, these methods may not be applicable to all types of models, and their performance can be limited.

Model-agnostic methods, on the other hand, focus on developing techniques that can be applied to any type of model to enhance its interpretability and transparency. These methods include feature selection, dimensionality reduction, and model compression. Feature selection is widely used method to identify the most relevant features for the model's predictions. By reducing the number of features, feature selection can enhance the model's interpretability, reduce the risk of overfitting, and improve its performance.

In this paper, we focus on the problem of enhancing the interpretability of EBMs by using cross features selectors and ensemble feature selection methods. By combining diverse feature selection methods, the ensemble can identify a set of features that are both relevant and complementary, reducing the risk of spurious interactions and single-feature dominance.

## II. Related Work

Hoque et al. (2018), noted many filter-based feature selection methods had issues of: not considering the redundancy among the selected features; having biasness on selected feature subset and inconsistent prediction accuracy during classification. To overcome these problems, the authors test multiple feature selection methods to select an optimal subset of non-redundant and relevant features. In their ensemble-based feature selection method, multiple feature subsets are combined to select an optimal subset of features using combination of feature ranking that improves classification accuracy. The authors were able to conclude from their average classification accuracy analysis, that the proposed ensemble mostly overcomes the local optimal problem of the individual filters especially for high dimensional datasets.

Pes (2020) explored ensembles of filter-based univariate and multivariate feature selection methods by using a simple ranking approach. The author tested their ensemble of 5 filter methods across 18 datasets for SVM and Random Forest and eventually concluded ensembles significantly improves the robustness of the selection process with no degradation in the predictive performance of the selected subsets. Binbusayyis et al. (2019) combined features of four filtered which occurred more than three times to form a new subset of features. This subset gave similar performance to individual selectors but was deemed to be more robust.

Thimoteo et al. (2022) compared the interpretability for white blood cell features and the presence of other pathogens in COVID-19 detection. The author reports how different models have varying feature importance and SHAP provided the most intuitive results. Yang et al. (2013) study instability across feature selection method and ensembles. The authors explored stability by varying sample size, sample order, data partitioning across filter and wrapper feature selectors. Furthermore, the authors explore stability of ensembles by varying data perturbation techniques, data partitions. The authors conclude that overall ensembles are largely robust to various data manipulations and processing techniques.

Effrosynidis et al. (2021) experimented with 12 individual feature selection methods, covering all three of the basic categories, i.e., filter, wrapper, and embedded, and 6 ensemble approaches. Their ensembles included Borda Count, which is the aggregation of individual ranks, and complex ensembles derived from the voting domain, such as Condorcet, Coombs, Bucklin, Instant Runoff, and Reciprocal Ranking. Their experimental results show that Reciprocal Rank outperforms all other methods in performance and has a high stability.

Lundberg et al. (2018) identifies that several common feature attribution methods for tree ensembles are inconsistent, meaning they can lower a feature's assigned importance when the true impact of that feature actually increases. This can prevent the meaningful comparison of feature attribution values. Further, they conclude that SHAP values are more stable and better align with human intuition. influential features. They finally present an extension of SHAP to measure hidden pairwise interaction relationships.

Nori et al. (2019) introduced Explainable Boosting Machine (EBM) which is a tree-based, cyclic gradient boosting Generalized Additive Model with automatic interaction detection. The boosting procedure is carefully restricted to train on one feature at a time in round-robin fashion using a very low learning rate so that feature order does not matter. And the round-robin cycles through features to mitigate the effects of co-linearity and to learn the best feature function for each feature to show how each feature contributes to the model's prediction for the problem. Second, EBM can automatically detect and include pairwise interaction terms.

The following are the issues that we observed in literature:
- EBM's is relatively a newer variant of boosting machines and is largely unexplored. There is almost no literature discussing the irregularities and challenges with EBM.
- Although "black box explainers" is a well-researched topic, there are only a handful literature available which compare the differences in outputs across the various approaches.
- Feature Selection methods are largely restricted to filter, wrapper and embedded methods of feature selections.
- Multi-step approach of feature selection although largely applied in industry has not be adequately discussed in Literature.
- Most Feature importance modules are used overlooking their deficiencies with the notion "It is what it is" and there has been very few explorations in improving these modules.

With these observations as our motivation, we identify two challenges with use of EBM's i.e. spurious feature interactions with redundant variables and Single feature dominance in interactions by using a multi-step cross feature selection to extract meaningful features to avoid interactions with redundant features and prevent a single feature dominance.

### III. Proposed Methodology

In this section, we will explain the proposed approaches experimented to eliminate spurious feature interactions and Single feature dominance in EBM.

**A) Ensemble 1:**

In this approach, we reduce the feature space by using a first stage cross-feature selector of a different algorithm. This new feature selector will eliminate redundant features which will not be part of the final EBMs modelling and thereby not allowing the EBM feature selector to interact with reductant features and find patterns. The Pseudo code is as depicted as below

i. Input: data X with features and n instances.
ii. Perform basic pre-processing of data
iii. for Feature Selectors→ $F_k$ in 0…. K selectors
   a. Set specific paraments for $F_k$
   b. Use a $F_k$ to select a subset of features, X', from X. Let X' have p features, where p ≤ m.
iv. Feed X' to Explainable Boosting Machine (EBM) model, f(X'), and obtain the predictions y'.
v. Output: Model performance and feature importance's.

We propose the use of a variety of state-of-the-art feature selectors $F_k$ as the first-round cross-feature selectors. They are SHAP, ADABOOST, XGBoost, Random Forest, Correlation, VIF, Variance Threshold, Permutation Importance and Boruta. A brief introduction for the cross-Features selectors are as follows:

1) SHAP (Lundberg et al. 2017): SHAP is a popular method for computing feature importance scores, which measures the contribution of each feature to the prediction of a model. The SHAP value for a feature represents the average marginal contribution of the feature across all possible coalitions of features. The SHAP values can be used to rank features and select the most important ones for a given model.

2) AdaBoost (Freund et al. 1997): AdaBoost is a popular ensemble learning method that combines multiple weak learners to form a strong learner. AdaBoost can also be used for feature selection by assigning weights to the features based on their importance in the weak learners. Features with high weights are more likely to be important predictors of the target variable and can be selected for the final model.

3) XGBoost (Chen et al. 2016): XGBoost is a popular gradient boosting method that uses a tree-based ensemble to predict the target variable. XGBoost provides feature importance scores based on the gain, cover, and frequency of each feature in the decision trees. Features with high scores can be selected for the final model.

4) Random Forest: Random Forest uses Mean Decrease in Impurity (MDI) to calculate top ranked features. MDI calculates each feature importance as the sum over the number of splits (across all trees) that include the feature, proportionally to the number of samples it splits.

5) Correlation: Correlation is a simple but effective method for feature selection that measures the linear relationship between features and the target variable. Features with high correlation coefficients are more likely to be important predictors of the target variable. Correlation can be computed using Pearson's correlation coefficient, which ranges from -1 to 1, or other variants such as Spearman's rank correlation.

6) VIF (Variance Inflation Factor): VIF is a measure of multicollinearity between features, which is a common problem in linear regression models. VIF computes the ratio of the variance of the estimated coefficients in a linear regression model to the variance of the same coefficients if the features were uncorrelated. Features with high VIF values are more likely to be redundant and can be removed to improve the model's performance.

7) Variance threshold: Variance threshold is a simple method for removing low-variance features, which are unlikely to have a significant impact on the model's performance. The variance threshold method computes the variance of each feature and removes those with variance below a certain threshold. This threshold can be set manually or automatically using techniques such as percentile or kurtosis-based thresholds.

8) Permutation Importance: Permutation Importance works on Mean Decrease in Accuracy (MDA). MDA is assessed for each feature by removing the association between that feature and the target. This is achieved by randomly permuting the values of the feature and measuring the resulting increase in error. The influence of the correlated features is also removed.

9) Boruta (Kursa et al. 2010): Boruta is a wrapper method for feature selection that uses a random forest model to evaluate the importance of features. Boruta works by comparing the importance of each feature with that of randomly generated shadow features. Features that are more important than the shadow features are selected for the final model.

A benchmark analysis is also conducted by using a plain EBM for feature selection. EBM is a fast implementation of the GA2M algorithm (Lou et al., 2013) and can automatically detect and include pairwise interaction terms of the form:

$$g(E[y]) = \beta_0 + \sum f_j(x_j) + \sum f_j(x_i x_j) \quad (1)$$

where, g is the link function that adapts the GAM to different settings such as regression or classification. The boosting procedure is carefully restricted to train on a single feature at a time in a round-robin fashion at a very low learning rate, so it doesn't matter how the features are ordered. To mitigate the effects of co-linearity and to learn the best feature function $f_j$ for each feature, it performs round-robin cycles through them to demonstrate how each feature contributes to the model prediction. Each function $f_j$ returns a term contribution and serves as a lookup table for each feature, enabling individual predictions to be made. To arrive at the final prediction, these term contributions are simply added together and processed by the link function g. In view of the modularity (additivity), term contribution can be arranged and assessed to show which highlights had the most effect on any individual prediction.

### B) Ensemble 2:

In this approach we experiment aggregating the features selected by the top first stage feature selectors by using (1)Features with Minimum overlaps across the Feature subsets and (2) merging all the subsets' top features. This new feature space will ensure to eliminate redundant features as wells as select the best of all features selectors which will form a new pool of features for the final EBMs predictions. The Pseudo code is as depicted as below

i. Input: data X with features and n instances.
ii. Perform basic pre-processing of data
iii. for Feature Selectors→ $F_k$ in 0.... K selectors
   a. Set specific paraments for $F_k$
   b. Create n feature pools using a feature selector, resulting in n subsets of features, F1, F2, ..., Fk.
iv. Aggregate the n feature pools to form two new feature pools using the following aggregation rules:
   a. Pool A: select the features that appear in at least k (k <= n) feature pools. Let FA be the subset of features selected for pool A. A value of K=4 is found to give most robust results I our experimentation.
   b. Pool B: select the top p (p <= m) features based on the aggregation scores. Let FB be the subset of top p features selected for pool B. A value of P=3 is found to give most robust results I our experimentation.
v. Obtain pools A and B: (X', FA) and (X', FB)
vi. Feed pools A and B to an Explainable Boosting Machine (EBM) model, f(X', FA, FB), and obtain the predictions y'.
vii. Output: Model performance and feature importance's.

### C) Altered EBM configuration:

In this approach EBM is used as a preselector to select the feature importance scores above the threshold 'M'. Next, we eliminate all interactions whose individual main feature importance does not rank better than the interaction. We next use the final pool of the individual features for the final feature selection in EBM. Based on experimentation for robust results the final value of M is 0.05.

### IV. EXPERIMENTAL SETUP

A. DATASET:
Three Benchmark datasets have been used to evaluate the performance of the experiments. The datasets are sourced from Kaggle, and the details are provided in Table 1

TABLE 1: SUMMARY OF DATASETS

| Dataset | Attributes | Data Instances | Class Distribution |
|---|---|---|---|
| Credit Score Data[14] | 305 | 20000+ | 1:9 |
| Loan Approval Data[14] | 34 | 42000+ | 1:21 |
| High Credit Score Data[14] | 305 | 82000+ | 1:12 |

B. EVALUATION TASKS :
We evaluate the proposed setup across the following tasks:
1. EBM model benchmark analysis: Evaluation of a vanilla EBM model on the whole dataset and the top 20 features.
2. Cross Feature Selection analysis: Analysis of performance of the alternate feature selector combined to the EBM model as a first stage selector.
3. Feature Ensemble: evaluation of the pooled features from multiple selectors with the EBM model.
4. Configured EBM: Evaluation of EBM with thresholds to preselect features for the final EBM model.

C. EVALUATION METRICS:
The models are evaluated across the following 4 metrics:

A. F-1 score: F1 Score is a good performance measure as it seeks a balance between Positive predictive value and True positive rate in imbalanced dataset.

$$F1\ Score = \frac{2 \cdot True\ Positive}{2 \cdot True\ Positive + False\ positive + False\ positive} \quad (2)$$

b. ACCURACY: Classification accuracy is a naive metric. It is the proportion of correct predictions made by the model

$$Accuracy = \frac{True\ Positive + True\ Negative}{Total} \quad (3)$$

c. COMPUTATIONAL TIME: Computational time in seconds denotes the total time for computation of the final EBM model on the reduced dataset.

$$Speed = time\ for\ computation\ (sec) \quad (4)$$

d. SINGLE FEATURE DOMINANCE: Measures how frequently a feature occurs in top 5 interactions given the count is more than 1.

e. SPURIOUS INTERACTION: Measures how frequently a feature interacts with another noisy feature (in the bottom 10% percentile of feature importance scores) to form a combined importance in top 10 percentile.

TABLE 2: PERFORMANCE SUMMARY OF ALL FEATURE SELECTORS

| | Dataset 1 | | | | Dataset 2 | | | | Dataset 3 | | | |
|---|---|---|---|---|---|---|---|---|---|---|---|---|
| | # Feat | F1 | Accuracy | Time (sec) | # Feat | F1 | Accuracy | Time (sec) | # Feat | F1 | Accuracy | Time (sec) |
| Plain EBM (whole dataset) | 34 | 99.93 | 99.89 | 12 | 305 | 97.79 | 96.01 | 21 | 34 | 97.16 | 96.69 | 69 |
| Plain EBM (Post Feature Selection dataset) | 20 | 99.94 | 99.89 | 12 | 20 | 96.70 | 94.04 | 12 | 20 | 96.77 | 94.2 | 8 |
| SHAP | 20 | 99.96 | 99.93 | 18 | 20 | 98.31 | 96.97 | 14 | 20 | 97.99 | 96.38 | 10 |
| ADABOOST | 9 | 99.51 | 99.16 | 26 | 11 | 97.33 | 95.2 | 7 | 8 | 97.01 | 94.62 | 3 |
| XGBOOST | 13 | 99.97 | 99.95 | 6 | 6 | 96.98 | 94.58 | 4 | 6 | 96.75 | 94.17 | 5 |
| Random Forest | 8 | 99.96 | 99.93 | 26 | 12 | 96.79 | 94.23 | 7 | 11 | 96.54 | 93.77 | 6 |
| Correlation | 25 | 99.94 | 99.89 | 19 | 73 | 97.68 | 95.83 | 26 | 69 | 97.59 | 95.67 | 12 |
| VIF | 10 | 97.19 | 95.08 | 24 | 45 | 96.92 | 94.48 | 17 | 42 | 96.66 | 93.97 | 6 |
| Variance Threshold | 28 | 99.96 | 99.96 | 7 | 187 | 98.26 | 96.88 | 41 | 186 | 98.14 | 96.67 | 10 |
| Permutation Importance | 10 | 99.97 | 99.95 | 4 | 25 | 97.9 | 96.23 | 9 | 25 | 97.39 | 95.3 | 2 |
| Boruta | 12 | 99.95 | 99.95 | 6 | 25 | 98.12 | 96.28 | 8 | 25 | 97.25 | 95.3 | 3 |
| Ensemble 2- top N pooling | 8 | 99.97 | 99.95 | 4 | 8 | 97.89 | 96.22 | 8 | 8 | 97.61 | 95.72 | 8 |
| Ensemble 2- top N aggregation | 16 | 99.94 | 99.96 | 6 | 14 | 97.88 | 96.2 | 9 | 15 | 97.15 | 94.88 | 9 |
| Altered EBM | 14 | 99.95 | 99.90 | 12 | 20.01 | 96.71 | 94.05 | 12 | 20.01 | 96.78 | 94.21 | 8 |

TABLE 3: FEATURE DOMINANCE & SPURIOUS INTERACTIONS

| | Feature Dominance | | | Spurious Interactions | | |
|---|---|---|---|---|---|---|
| | Dataset1 | Dataset 2 | Dataset 3 | Dataset1 | Dataset 2 | Dataset 3 |
| Plain EBM (whole dataset) | 1 feature x 5 Occurrence | 3 feature x 2 Occurrence | 1 feature x 5 Occurrence | 1 | - | 2 |
| | | | | | | |
| SHAP | 1 feature x 3 Occurrence | 1 feature x 3 Occurrence | 1 feature x 3 Occurrence | | | |
| ADABOOST | 2 feature x 2 Occurrence | 2 feature x 2 Occurrence | 2 feature x 2 Occurrence | | | |

| | | | | | |
|---|---|---|---|---|---|
| **XGBOOST** | 1 feature x 2 Occurrence | 1 feature x 2 Occurrence | 1 feature x 2 Occurrence | | |
| **Random Forest** | 2 feature x 2 Occurrence | 2 feature x 2 Occurrence | 2 feature x 2 Occurrence | 1 | 1 |
| **Correlation** | 1 feature x 3 Occurrence | 1 feature x 2 Occurrence | 1 feature x 2 Occurrence | | |
| **VIF** | 1 feature x 2 Occurrence | 1 feature x 2 Occurrence | 1 feature x 2 Occurrence | | |
| **Variance Threshold** | 1 feature x 4 Occurrence | 1 feature x 2 Occurrence | 1 feature x 4 Occurrence | | |
| **Permutation Importance** | 2 feature x 2 Occurrence | 2 feature x 2 Occurrence | 2 feature x 2 Occurrence | | |
| **Boruta** | 1 feature x 5 Occurrence | 2 feature x 2 Occurrence | 1 feature x 5 Occurrence | | |
| **Ensemble 2- top N pooling** | 1 feature x 2 Occurrence | 1 feature x 2 Occurrence | 1 feature x 2 Occurrence | | |
| **Ensemble 2- top N aggregation** | 2 feature x 2 Occurrence | 2 feature x 2 Occurrence | 2 feature x 2 Occurrence | | |
| **Altered EBM** | 1 feature x 3 Occurrence | 2 feature x 2 Occurrence | 1 feature x 3 Occurrence | | |

## V. RESULTS AND ANALYSIS

The experimental results were conducted on a machine with a memory of 16GB with an Intel core i5-8250U CPU@2.60GHZ X 8 processor with an INTEL UHD Graphics 620 with GNOME 3.28.2 and 64-bit OS with a disk of 320 GB. The programs are written in Python 3.7.3 with the help of various libraries like Sklearn, NumPy, StatsModel etc. The three datasets are preprocessed by(i) imputing missing values by -9999, (ii)categorical feature encoding and (iii)feature scaling. The training and testing split is in the ratio 70:30. The cutoff threshold for correlation is 0.7 & 0.02 is the feature importance cutoff for all other selectors.

PERFORMANCE :

The performance summary of all the feature selectors in comparison with vanilla and finetuned EBM is reported in Table 2. It is very evident from the results that a multistage feature selector outperforms the finetuned EBM with 20 features across all experiments.

The better performance is largely accredited to the power of ensembles and reduction of noise the pre-feature selectors. Overall SHAP, Boosting, Correlation, Variance Thresholds and ensemble preselectors is found to be a good choice a preselector for EBM's. Different pre-feature selectors bring different value to the process, for instance SHAP values learn to map relations of individual contribution by game theory approaches unlike other approaches. While, XGBBOOST has a regularization term which helps penalize non-important features and eliminate from the feature selection. Thus, each method has their own advantage which help reduce the sample space. EBM's however wont be able to reduce the sample space as effectively as they tend to select the top features based on main contribution and interaction contributions, thus allowing redundant features to be retained.

FEATURE DOMINANCE & SPURIOUS INTERACTIONS :

The summary of Feature Dominance and Spurious Interactions eliminated is reported in Table 3. A top-level overview clearly shows that by using a cross feature selector prior to EBM's both feature dominance and spurious Interactions are largely eliminated. To bring in generalization of results we have experimented on 3 different datasets so that the results are not data specific. It can be noted that most interactions modelled by EBM's are between highly correlated and redundant variables. However, the elimination of multicollinearity and redundancy by the pre-feature selectors clearly reduces noise from the feature space thereby limiting EBM to interact with only non-trivial features. Another major reason is the fact ensemble with a pre-selector ensures missing features and outliers are adequately removed rather than being inherently handled thereby making predictions more meaningful.

Table 4 provides the top 5 interactions for Dataset 1 using EBM. The feature "recoveries" occur in all 5 interactions and is clearly dominating, correlated and prohibits meaningful interpretations. Table 5 reports the improved interactions by using XGBoost Preselector prior to EBM. The feature interactions are no more dominated by the feature "recoveries". This is largely due to having a limited feature space of non-trivial features. This limited feature space lacks noise or any redundancy which coincidentally had some predictive power when interacting with a highly predictive variable. Finally the Interaction are limited to highly important variables which also bring intuitive reasoning on its contribution.

TABLE 4: FEATURE DOMINANCE IN EBM (TOP 5 INTERACTIONS)

| Feature Interaction | Importance Score |
|---|---|
| 12 ('recoveries x Month_last_pymnt_d' | 0.40439605400365486) |
| 17 ('recoveries x last_pymnt_amnt' | 0.29194822925563263) |
| 18 ('recoveries x Month_last_credit' | 0.25982373253753716) |
| 20 ('recoveries x Year_last_pymnt_d' | 0.2381425414885942) |

| Feature Interaction | Importance Score |
|---|---|
| 21 ('loan_amnt x recoveries' | 0.229985082930677306) |

TABLE 5: FEATURE DOMINANCE ELIMINATED IN XGBOOST + EBM (TOP 5 INTERACTIONS)

| Feature Interaction | Importance Score |
|---|---|
| 11 ('recoveries x total_rec_prncp' | 0.20362923770336663) |
| 13 ('total_rec_prncp x loan_amnt' | 0.16224080140105795) |
| 15 ('60_months x last_pymnt_amnt' | 0.14342356268711523) |
| 17 ('Year_last_credit x loan_amnt' | 0.07025993923083294) |
| 18 ('last_pymnt_amnt x loan_dur' | 0.058044544036164616) |

## VI. CONCLUSION

Based on the experiments conducted it can be concluded that using a multistage cross feature selection process not only improves the performance of EBM's but also can bring feature stability and remove spurious interaction with noisy features. Results show that SHAP, Boosting, Correlation, variance Thresholds and ensemble preselectors convincingly out beat the pairwise interaction-based feature selection of EBM's on noisy datasets. Furthermore, there is considerable decrease of feature dominance as only the meaningful interactions remain after removal of redundant features. Also, it may be noted that at all the spurious interactions were in fact not important for the model prediction and removal of redundant features improved the interaction interpretability.

The challenge to this research is the plethora of choices with respect to methods, algorithms and datasets available. It is hard to experiment all available techniques and datasets to prove generalizations. This research was limited to two major issues of EBMs and the experiments only highlighted few industry prevalent techniques to handle redundancy and improve performance. For future work, there are several potential avenues for future work. One potential avenue is to further investigate why certain feature selectors outperform other methods. Another potential direction would be to explore other state of the art Explainable algorithms like Gami-Net (Yang et al. 2021) for their stability. Finally, one can experiments for methods to determine generalization and effectiveness in feature selection processes.